# Are Data Augmentation and Segmentation Always Necessary? Insights from COVID-19 X-Rays and a Methodology Thereof

Aman Swaraj[1, *, +], Arnav Agarwal[1,2, +], Hitendra Singh Bhadouria[3,4], Sandeep Kumar[1, *], Karan Verma[4]

aman_s@cs.iitr.ac.in[1], arnav.agarwal@gmail.com[2], 171210028@nitdelhi.ac.in[4], sandeep.garg@cs.iitr.ac.in[1], karanverma@nitdelhi.ac.in[4]

[1] Department of Computer Science and Engineering, Indian Institute of Technology Roorkee, India – 247667
[2] Department of Computer Science and Engineering, Manipal University, Jaipur, Rajasthan, India – 302034
[3] United Health Group, Optum division
[4] Department of Computer Science and Engineering, National Institute of Technology, Delhi, India – 110036

[+] these authors contribute equally to the work

**\*Corresponding Author:**
Aman Swaraj, Sandeep Kumar

Institute Address:
Department of Computer Science and Engineering,
Indian Institute of Technology Roorkee,

Roorkee, Uttarakhand,
India – 247667

Tel: 91-7217842795; 91-7579024426
Email: aman_s@cs.iitr.ac.in; sandeep.garg@cs.iitr.ac.in

**Word count excluding tables, figures, captions and references = 5030**

**ABSTRACT**

**Purpose**: Rapid and reliable diagnostic tools are crucial for managing respiratory diseases like COVID-19, where chest X-ray analysis coupled with artificial intelligence techniques has proven invaluable. However, most existing works on X-ray images have not considered lung segmentation, raising concerns about their reliability. Additionally, some have employed disproportionate and impractical augmentation techniques, making models less generalized and prone to overfitting. This study presents a critical analysis of both issues and proposes a methodology (SDL-COVID) for more reliable classification of chest X-rays for COVID-19 detection.

**Methods**: We use class activation mapping to obtain a visual understanding of the predictions made by Convolutional Neural Networks (CNNs), validating the necessity of lung segmentation. To analyze the effect of data augmentation, deep learning models are implemented on two levels: one for an augmented dataset and another for a non-augmented dataset.

**Results**: Careful analysis of X-ray images and their corresponding heat maps under expert medical supervision reveals that lung segmentation is necessary for accurate COVID-19 prediction. Regarding data augmentation, test accuracy significantly drops beyond a certain threshold with additional augmented images, indicating model overfitting.

**Conclusion**: Our proposed methodology, SDL-COVID, achieves a precision of 95.21% and a lower false negative rate, ensuring its reliability for COVID-19 detection using chest X-rays.

**Index Terms:** COVID-19, Chest X-ray, Lungs segmentation, Class Activation Mapping, Data Augmentation

# 1 INTRODUCTION AND RELATED WORKS

The initial cases of Covid-19 appeared in early December 2019 in Wuhan, China. Although the identified pathogen shared a phylogenetic similarity to SARS and MERS, it has caused much more causality than its preceding variants. Attributed as the first pandemic of the 21st century, Covid-19 has already infected around 700 million people and caused more than 7 million deaths in last 2.5 years.

In the initial days of the spread, most of the countries, due to unavailability of any vaccine, focused on curbing the virus through lockdowns, social distancing measures and forecasting models [1]. However, almost a year after the reporting of first case in Wuhan, WHO listed the Comirnaty COVID-19 mRNA vaccine for emergency use, making the Pfizer/BioNTech vaccine the first to receive emergency validation. Since then, almost all countries have instigated vaccination drives to ensure safety of their citizen. However, despite safety protocols and widespread vaccinations, the COVID-19 virus has continued to mutate, producing variants such as Delta and Omicron, the latter being highly transmissible. These mutations have posed significant challenges, particularly in developing countries with limited resources and high population densities, where even minor discrepancies in strategies can result in large-scale casualties.

While COVID-19 is receding as a global threat, the emergence of other respiratory illnesses, such as the recent human metapneumovirus (HMPV) outbreak in China[1], emphasizes the need for vigilance against future pandemics. These scenarios underline the importance of developing robust diagnostic tools capable of rapid and accurate disease detection to prevent widespread casualties.

---

[1]https://www.who.int/emergencies/disease-outbreak-news/item/2025-DON550

One widely used clinical technique for diagnosing covid-19 is the real-time reverse transcription-polymerase chain reaction (RT-PCR). However, the reported accuracy of RT-PCR falls around 60% and thus doctors have successively prescribed chest X-rays for better diagnosis [2].

Artificial Intelligence (AI) has played a key role in recent advancement of biomedical research. Deep neural networks, especially, Convolutional Neural Networks (CNN) have emerged quite successful in the domain of medical imaging [3]. These techniques carry the potential to help radiologists diagnose covid-19 patients by screening the chest X-rays. Subsequently, many researchers have used deep learning models for classifying covid-19 infected lungs.

Since covid-19 is a recent disease, the dataset of covid-19 infected lungs are relatively scarce and therefore many researchers have applied data augmentation techniques to increase the training data. In many cases, authors have used techniques such as scaling, flipping, shear, zoom, and rotation of X-ray images to increase the dataset. However, the usage of certain methods such as horizontal flipping is open to question, since, during testing or real-world usage, the X-Rays would never be flipped. Some authors have also used Generative Adversarial Networks (GAN) to generate dataset, but again, the reliability of GAN in such a use case needs to be validated. Above all, the proportion in which the data has been generated can very much lead to over fitting and loss of generalization of the model. Moujahid et al. also point out this in their study [4] that the obtained results by augmentation techniques are not efficient to be projected for the real world.

For instance, Islam et al. [5] trained their model on a total of 1220 COVID-19 images. However, out of those 1220 images, 912 images were augmented. Khalifa et al. [6] used GAN for data augmentation and increased the dataset 10 folds which is again not reasonable. In [7], Ahmed et al. used 472 images to generate 2000 images. Rahman et al. [8] also generated twice the number of initial images. Similarly, authors in [9-14] also used data augmentation techniques. Much recently, Masadeh et al. [15] also performed data augmentation to increase their training dataset which initially comprised of only 217 covid images. One common observation about most of the earlier works is that, by default, they have assumed augmentation to improve the performance. However, a comparative analysis of models trained on un-augmented and augmented dataset is still lacking.

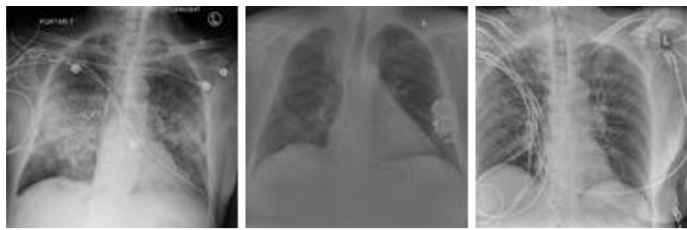

Figure 1: Presence of pacemakers and ECG wires in X-Ray images (source- Dataset [21])

Besides data augmentation, many studies have not considered segmentation of lungs before training the models which again makes the model unreliable as without lung segmentation, the model can learn wrong features from outside the lungs portion. Pertinently, Gianluca Maguolo et al. in their study [16], highlighted how even after removal of lungs from the x-ray images, the neural network still gave good results which implies that the model is also learning features from outside of lungs. This is a significant point as features outside the lungs are irrelevant in case of covid detection and could lead to misclassification. Even if somehow, they classify correctly, the model's reliability would still remain questionable. From fig 1, it is very much evident that, besides the presence of chest and collar bones, the x-rays also contain additional elements such as ECG wires, pacemakers etc. from which the neural network can extract features. Rahman et al [8] also showed in their work how non-segmented x-rays are unreliable. Hitendra et.al. [17] in their work point out that on an estimate 30% of the covid positive images have this anomaly while it is absent in the x-ray images of covid negative patients, thus making the model highly unreliable. Similar concern for validity of the deep learning models are shown by authors in [18, 19]. In this work, we further apply Grad-CAM to validate the usage of lungs segmentation. We further apply Grad-CAM in our current work to validate this claim.

Although some authors have applied segmentation, we would like to point out certain trade-offs in their work. For instance, Tabik et al. [20] in their work on severity based covid classification, used segmentation-based cropping to improve the sensitivity of their results. However, their accuracy dropped drastically for cases with moderate or mild symptoms. From utility perspective, while they have inferred their model on different levels of severity and presented the accuracy, it doesn't have any practical utility in that direction as in real time scenario, a patient with severe symptom could be identified even without the use of X-ray as their symptoms would be visible externally. However, a patient with mild infection whose symptoms are not much visible outwardly, needs to be analyzed via X-rays. Also, the actual accuracy from a practical standpoint should come by taking average of their accuracies for different severity. Further their study is based on smart data, i.e., there data is carefully selected to have best features. Again, this is impractical in real time scenario. Rather, theoretically, a larger heterogeneous data sample should make the model more generalized.

Similarly, Teixeira et al. [13] also applied segmentation techniques. However, they could only achieve a highest f-1 score of 0.83. Ahmed et al. [7] achieved good results after applying lungs segmentation. However, their test size was very small and therefore loss of generalization can happen. They also used data augmentation techniques. Similarly, Rahman et al [8] also generated twice the number of original covid 19 images. This level of augmentation might lead to overfitting and does not really offer multiple datapoints.

Another pertinent work in this direction is carried out by [21] where they validate the usage of lungs segmentation, but their work is limited only to the validation part and they don't provide any new model based on their observations. Further, their data is highly skewed and they also augment images to a high degree. Nevertheless, their data agrees with importance of segmentation and we further validate the same on a more balanced and coherent dataset. A brief overview of some of the major studies along with their drawbacks is listed in table 1.

Concerning the points stated above, it can be concluded that a much-detailed study of the impact of lungs segmentation and data augmentation on covid-19 data is needed to better understand the nuances involved in covid-19 classification.

With this motivation, we attempt to answer two research questions (RQs) in this study:

- **RQ1. Is Lungs Segmentation really necessary for accurate classification of covid-19 images?** AI based medical imaging can help in making clinical decisions. At the same time, they are highly volatile in nature and therefore, it is necessary to understand which features are contributing in the decision making. In case of covid-19, ideally, the decision should be made on the basis of infection in the lungs. However, a raw x-ray image can have features extrinsic to the lungs as well that could lead to decision making which is not desirable. Earlier works [19, 27, 28, and 30] have pointed out this possibility, but a more detailed analysis is required. We analyze the necessity of the segmentation of lungs in detail with class activation mapping and theoretical reasoning.

- **Does data augmentation necessarily improve the performance of the model?** Many studies have adopted data augmentation for increasing the dataset with an assumption that it will improve performance. Although, this assumption is based on standard practices or theoretical backing, yet, if not properly analyzed, it can lead to loss of generalization and over fitting. We aim to inspect the effect of the same and further, use of certain augmentation methods such as horizontal flipping and rotation also needs to be checked.

Table 1: Brief overview of existing work on covid-19 lungs classification

| Reference | Limitations |
|---|---|
| Islam et al. [5] | • Lungs segmentation absent – therefore unreliable.<br>• 75% Data is augmented – therefore chances of overfitting. |
| Khalifa et al. [6] | • Lung segmentation absent.<br>• 90% data generated via augmentation – could lead to overfitting. |
| Ahmed et al. [7] | • 80% data augmented- could lead to overfitting.<br>• Final test size is very low |
| Rahman et al. [8] | • 50% data augmented - could lead to loss of generalization. |
| Moris et al. [9] | • Very small dataset with total 720 images and only 240 covid images.<br>• They have used GAN. The reliability of GAN is questionable in this scenario.<br>• Lung Segmentation missing. |
| Wang et al. [10] | • Used horizontal flipping for generating new data which is invalid as x-ray images would never be flipped in real time scenario.<br>• Lung Segmentation absent. |
| Narin et al. [11] | • Dataset size very low<br>• Lungs Segmentation absent.<br>• Uses Horizontal flipping for augmentation – Not Valid |
| Abed et al. [12] | • Used horizontal flipping- Not Valid<br>• They generate 24000 images from 400 original images which is 60x augmentation-very high loss of generalization<br>• Lungs segmentation absent<br>• Qualitative analysis missing |
| Teixeira et al. [13] | • Imbalanced class distribution<br>• Low performance on segmented dataset.<br>• Used horizontal flipping – not valid. |
| Waheed et al. [14] | • Lungs segmentation absent<br>• Used GAN for increasing dataset which is again questionable. |
| Masadeh et al. [15] | • Lungs segmentation absent.<br>• Used flipping for augmentation – Not Valid<br>• Covid images very less, only 217. |
| Tabik et al. [20] | • Impractical from utility point of view as accuracy drops drastically for patients having mild or moderate symptoms.<br>• Study is based on smart data, could lead to loss of generalization. |
| Sadre et al. [21] | • Investigation based work, no new model is proposed based upon the observation.<br>• Highly skewed data.<br>• Degree of data augmentation is also quite high. |
| Fang et al. [22] | • Small dataset, low accuracy |
| Yang et al. [23] | • Lungs segmentation is absent.<br>• Grad cam analysis is also not carried out. |
| Tang et al. [24] | • Lungs segmentation absent.<br>• Test dataset very small. |

| Reference | Limitations |
|---|---|
| Ucar et al [25] | • Lungs segmentation absent. |
| | • Used flipping for data augmentation which is not valid. |

Based on the findings of the above two RQs, we then propose our methodology, SDL-COVID for a much reliable and accurate classification of covid-19 x-rays.

In summary, following are the contributions of this work:

1. We present a methodology, SDL-COVID for covid-19 x-ray classification that includes lungs segmentation which is absent in most of the earlier works.
2. We perform class activation mapping on the dataset to validate the usage of lungs segmentation along with theoretical reasoning.
3. Earlier works have directly assumed that data augmentation improves performance. In our work, we train the model with and without augmentation and inspect the results to verify whether augmentation actually helps or not.
4. Further, we also present the reasoning why certain augmentation techniques are impractical and how disproportionate augmentation can lead to overfitting and loss of generalization.

Rest of the paper is organized as follows: In section 2, we describe the dataset source and discuss all the methods carried out in the study. Section 3 presents the experiment results. Section 4 holds a discussion and finally, in section 5 we conclude with future work.

## 2 METHODOLOGY

We carry out our research on three levels, first we discuss the two RQs, followed by the proposed methodology. We start with the first research question to validate the usage of lungs segmentation (fig. 2). We apply class activation mapping on the x-ray images to visualize the feature extraction process. Then for the second question, we run the models on two levels, one for augmented dataset, and another on non-augmented ones. Fig. 3 then shows the steps involved in RQ2 concerning data augmentation. Based on the findings of the first two RQs, we then propose our methodology SDL-COVID as presented in fig. 4 and Algorithm 1. Image enhancement techniques applied on the dataset are mentioned in section 2.1. Section 2.2 describes the U-net architecture [35] used for segmentation. Section 2.3 briefly describes the features extraction techniques used in our study. Finally, in section 2.4, we talk about the performance metrics.

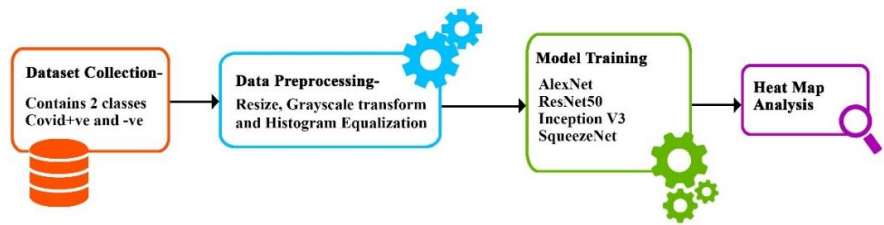

Figure 2: Steps involving RQ1 for validation of usage of lungs segmentation.

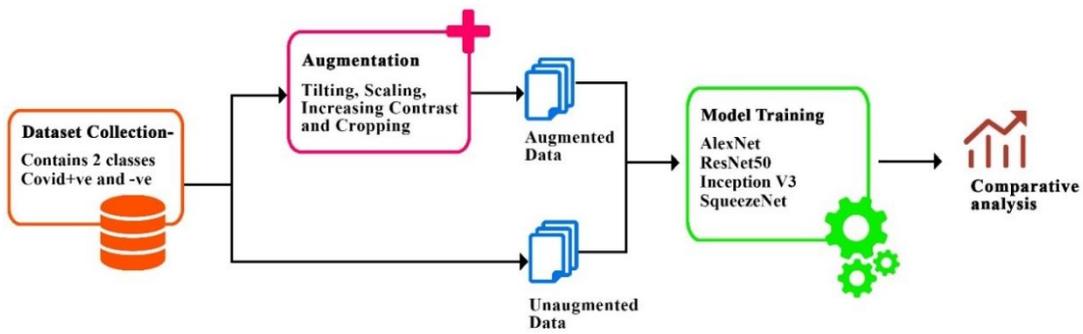

Figure 3: Steps involved in RQ2 concerning data augmentation.

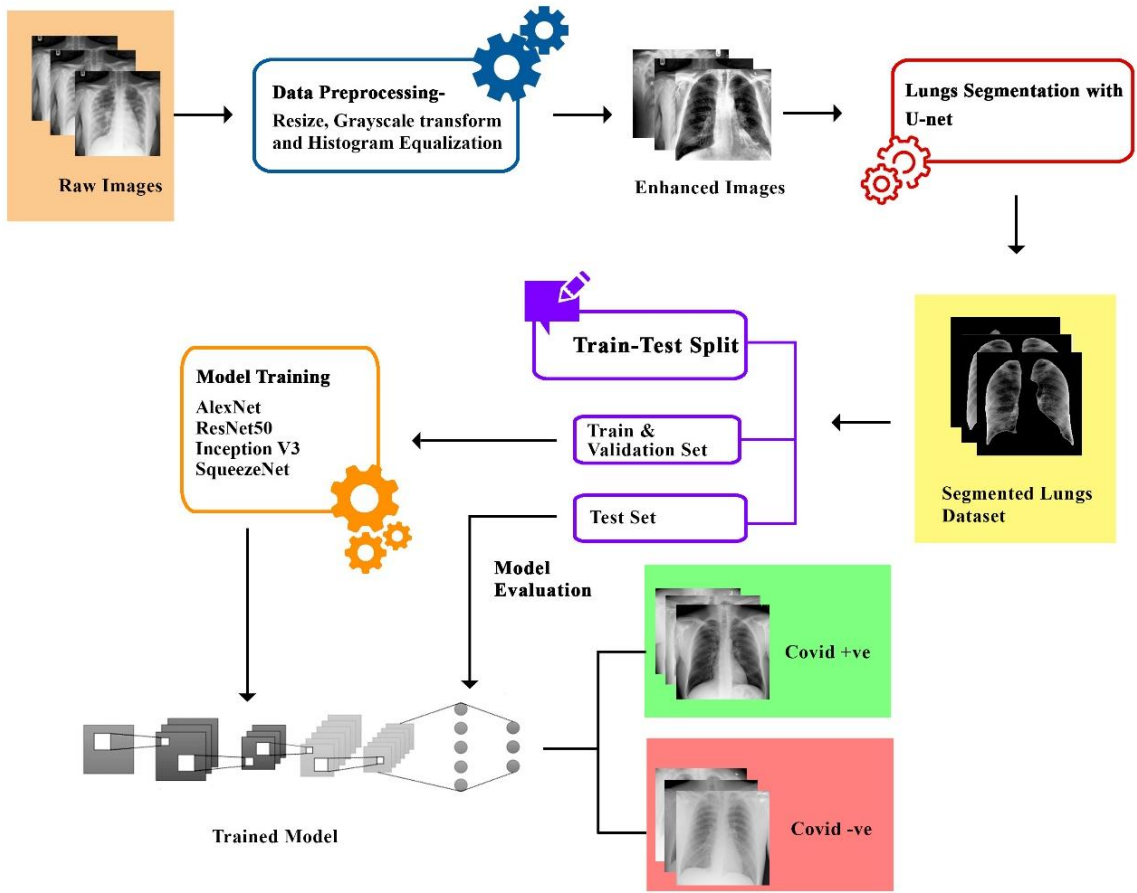

Figure 4: Our proposed approach, SDL-COVID.

ALGORITHM 1: Pseudo Code of the proposed methodology, SDL-COVID.

**Input:**

$N$: the size of CXR images dataset; $D_i$:

$i^{th}$ input sample of CXR images dataset $D$;

$S_i$: the pre-processed and segmented image corresponding to $D_i$;

$T_i$: True class of input image $D_i$;

$M$: Set of compiled CNN Models {AlexNet, ResNet50, InceptionV3, SqueezeNet}.

**Procedures:**

*HistEq (i)*: applies histogram equalization to input image *i*.

*Gray (i)*: applies grayscale to input image *i*.

*Segment (i)*: applies lung segmentation to image i using pretrained U-Net model.

*Split (N)*: distributes integers 1 to *N* into three subsets train (65%), validation (15%), test (20%).

*Train (m, d, t, v)*: trains model *m* using on dataset d using training split *t* and validation split *v*.

*Predict (m, i)*: predicts class using model *m* on input image *i*.

*Evaluate (TP, FP, TN, FN):* calculates and prints various evaluation metrics using count of true positives, false positives, true negatives and false positives.

**Output:**

*Pm, i:* Predicted Class for input image *i* using model *m*.

**for** *i* = 1 to *N* **do**

    *t* = Gray ($D_i$)

    *t* = HistEq(*t*)

    $S_i$ = Segment(*t*)

**end for**

*train, val, test = Split(N)*

**for** ¥ *m* € *M* **do**

    Train (*m, S, train, val*)

    **for** ¥ *i* € *test* **do**

        $P_{m,i}$ = Predict (*m, $S_i$*)

    **end for**

**end for**

**for** ¥ *m* € *M* **do**

    TP = 0

    TN = 0

    FP = 0

    FN = 0

    **for** ¥ *i* € *test* **do**

        **if** $T_i$ == $P_{m,i}$ **and** $T_i$ == 'covid' **then**

            TP = TP + 1

```
        else if T_i == Pm, i and T_i == 'normal' then
                TN = TN + 1
        else if T_i != Pm, i and T_i == 'normal' then
                FP = FP + 1
        else
                FN = FN + 1
    end for
    Evaluate (TP, FP, TN, FN)
end for
```

---

## 2.1 Image Enhancement

Since, raw images are of varying features in terms of image resolution, color, shape and size, so, in order to standardize the images in the dataset, all images were converted into grayscale color format and the dimensions changed to those suited best for the various models, i.e., 244x244 for Alexnet and ResNet50, 299x299 for Inceptionv3, and 224x224 for Squeezenet respectively. Further to enhance the features required for classification of covid infection, appropriate image enhancement techniques are applied. Zhao et al. [27] in their study noted that confirmed patients had typical imaging features that could be helpful in identifying other suspected patients. Ground-glass opacities (GGO) (87 [86.1%]) or mixed GGO and consolidation (65 [64.4%]), were among the imaging features mentioned in the study. In simpler terms, GGO refers to a hazy lung opacity having a thin veil like appearance on the X-ray image being not able to obscure underlying bronchial walls or pulmonary vessels. Appearance of GGO are vividly described in figure 5.

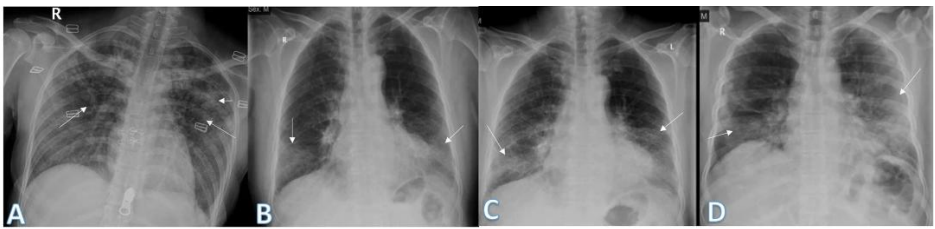

Figure 5: Symptoms of Covid-19 infection evident in x-ray images- (a) bilateral upper zonal ground glass opacities (long arrows) with small left upper zonal air space consolidation opacities (short arrow). (b-c) Primary and secondary chest X-ray revealing bilateral lower zonal ground glass opacities (arrows). (d) Initial chest X-ray showing right lower zonal and left mid ground glass opacities (arrows) [28].

In order to accentuate these features corresponding to the infection, we explored various image enhancement techniques, such as, Gaussian Unsharp mask, Laplacian Unsharp Filter, Butterworth low pass filter, Histogram equalization (HE), and Contrast Limited Adaptive Histogram Equalization (CLAHE). The main aim of applying these filters was to make prominent the imaging features that are indicators of Covid-19 infection, such as GGO. Out of all the filters, HE and CLAHE showed the greatest improvement in clarity, however, CLAHE greatly increased the visibility of bones and wires as compared to HE. Hence, for the image enhancement, we decided to use HE. Fig 6 shows various image enhancement techniques applied on few x-ray images.

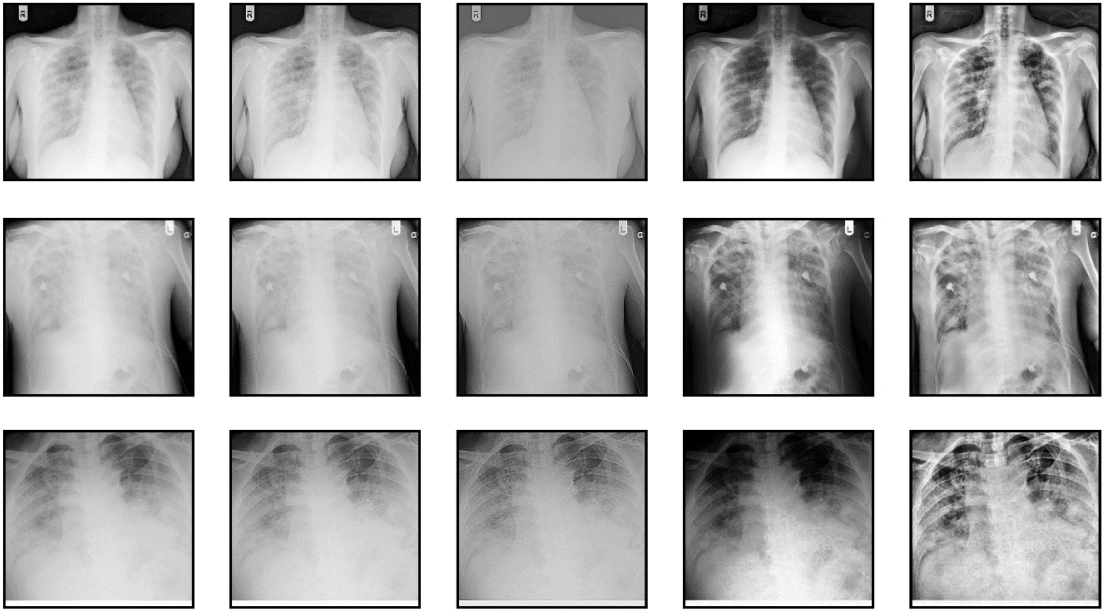

Figure 6: From left – Original, Gaussian Unsharp Mask, Laplacian Unsharp Mask, Histogram Equalization and CLAHE.

## 2.2 Lungs segmentation

As per our hypothesis, we apply CAM to validate the usage of lungs segmentation. Post validation, we use U-Net architecture to segment the lungs. U-Net is a fully convolutional neural network adept for semantic segmentation in biomedical images. Fig.7 shows the resultant images after segmentation.

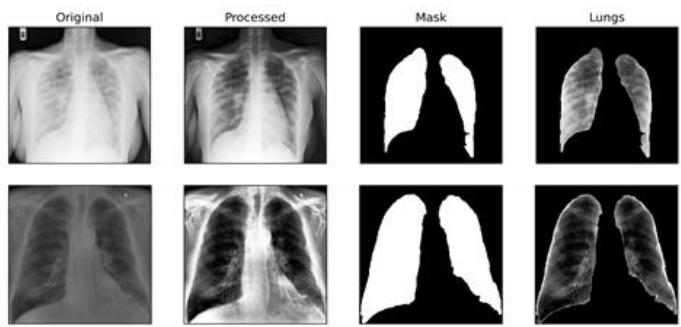

Figure 7: From left- Original X-ray, Enhanced X-ray, Lung segmentation Mask, Segmented Lungs.

## 2.3 Feature extraction models

For this project we explored numerous images classifying models that were available on the PyTorch repository, we narrowed down on four models i.e., ResNet50 [29], InceptionV3 [30], AlexNet [31], and SqueezeNet [32]. Further, we employed the keras library from TensorFlow for Grad-CAM implementation. A heat map was created by computing the gradient of the top predicted class (of the input image) using the activations of the last convolution layer. A vector was then defined, which was the mean intensity of the gradient of each feature map. Since feature maps are the output channels created from convolutions, they help us visualize the learning process in CNNs and show the most valuable features, of an input, the CNN used to arrive at a prediction. Each feature map contains new weights (or "importance") values. We multiply these values of each such channel in the feature map array, with regard to the top predicted class. The channel-wise mean of this resulting feature map is our heat map of class activation.

## 2.4 Performance Evaluation Metrics

We use Accuracy (ACC), F1-Score (F1), Precision (Pre), Sensitivity (Se), and Specificity (Sp) as primary performance metrics along with negative prediction value, false positive rate, false discovery rate, and false negative rate.

Accuracy is a measure of the total number of correct predictions made by the model on the entire test set. F1-Score is essentially the harmonic mean of the model's precision and sensitivity. Precision deals with how well the positive predictions have been carried out by the model.

Given a patient has symptoms, the probability that the test results are also positive is defined by sensitivity. Specificity on the other hand is the probability of a negative test result given the condition that the patient is not showing any symptom in real.

All these metrics are calculated using True Positive (TP), False Positive (FP), True Negative (TN), False Negative (FN) parameters generated from each class of the confusion matrix. Test results correctly indicating the presence of the symptoms accounts TP. While TN indicates the absence of the symptoms. FP indicate the cases where model has wrongly indicated the symptoms and FN points out those cases where the symptom is present, but not detected by the model.

$$Se = \frac{TP}{TP+FN} \quad (1)$$

$$Sp = \frac{TN}{FP+TN} \quad (2)$$

$$Acc = \frac{TP+TN}{TP+FN+TN+FP} \quad (3)$$

$$Pre = \frac{TP}{TP+FP} \quad (4)$$

$$F - Score = \frac{2*TP}{2*TP+FP+FN} \quad (5)$$

## 3 EXPERIMENT RESULTS

In this section, we discuss the outcome of the research experiments. Section 3.1 starts with dataset description and in 3.2, we analyze the feature extraction models. Section 3.3 and 3.4 elaborate on the RQs followed by the results of our proposed methodology in 3.5. we also make a comparative analysis with earlier existing works that have included segmentation in their work in section 3.6.

All the experiments were performed on Intel Core i5-9300H CPU, NVIDIA GeForce GTX 1660ti GPU, 16GB 2667 MHz RAM and Micron m.2 SSD.

### 3.1 Dataset Description

For our work, we use the publicly available dataset [10], 'COVIDx' which comprised of total 17742 posterior-to-anterior chest x-rays at the time of study. The dataset is an integration of images from multiple open-source chest radiography datasets having comparable demographics and age range. The images in the dataset are categorized as positive and negative x-rays having 2449 and 15168 images respectively for positive and negative cases of covid-19. Since the positive negative segregation of images have a huge imbalance, there are chances of over fitting, and loss of generalization which can lead to inaccurate results. To handle this, we randomly sample 2200 positive and 2100 negative x-rays from the dataset.

In order to check the impact of data augmentation (RQ2), we also incorporate the use of four different types of data augmentation strategies namely, tilting/skewing, scaling/changing aspect ratio, cropping, and changing contrast. For each augmentation method, from both positive and negative cases, we generate around 300 augmented images amounting to a total of 2400 augmented images.

### 3.2 Analyzing the CNN models

We selected four different CNN models for our experiments. Initially, we trained all the models on the original unsegmented data consisting of 2100 negative and 2199 positive images which were randomly sampled, and HE enhanced. The models were pre-trained on ImageNet [33]. We set the recommended mean as per standard values for all the models. The images in the dataset were also resized to dimensions that best fit with the model architecture.

On average, each model was trained for 30-40 epochs and reached a plateau in accuracy after 35+ epochs. When assessing model accuracies using non-segmented data, while AlexNet performed best in training with an accuracy of 99.0%, InceptionV3 excelled in testing with a 97.3% accuracy. However, when considering overall performance across all data types (including augmented and segmented data), SqueezeNet emerged as the top performer. It achieved the highest training accuracy at 98.5% and maintained a test accuracy of 96.4%. Additionally, it demonstrated competitive performance on augmented and segmented data compared to other models (Fig.8).

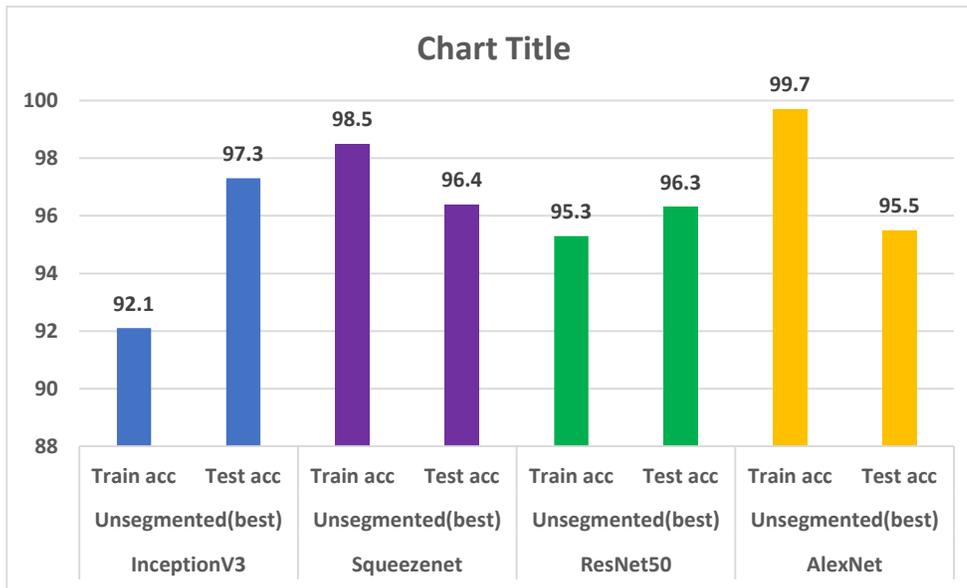

Figure 8: Performance of various CNN models on original dataset without segmentation.

### 3.3 RQ1. Is Lungs Segmentation really necessary for accurate classification of covid-19 images?

Covid-19 is a respiratory disease and therefore, decision should be made on the basis of infection in the lungs. However, a raw x-ray image also contains features extrinsic to the lungs portion that could contribute to decision making such as bones, pacemakers, ECG wires etc. as shown in fig 1. Given the volatile nature of medical imaging, this is not desirable. Earlier works [8, 16, 17, 21] have also highlighted this issue.

To validate our hypothesis, that neural networks can indeed learn features from outside the lungs area, we perform heat map analysis of the features extracted by the CNN models on the x-ray images. We start with basic pre-processing of the x-ray images followed by feature extraction. Finally, we take the help of the visual heat maps to identify the detection points of the CNN model and find out if they are medically reliable, by consulting with a qualified medical professional. Fig 9 shows some of the heatmaps over unsegmented lungs, more images can be found on our GitHub repository (https://github.com/Anonymous-sdlcovid/SDL-COVID).

From fig 9, it's quite evident that the network indeed is extracting features from outside the lungs area which is highlighted in yellow tone. Therefore, we can safely conclude that segmentation of lungs is indeed necessary in order to get authentic classification of covid infected lungs. In subsequent sections, we further explore the effect of segmentation on the decision making of the model.

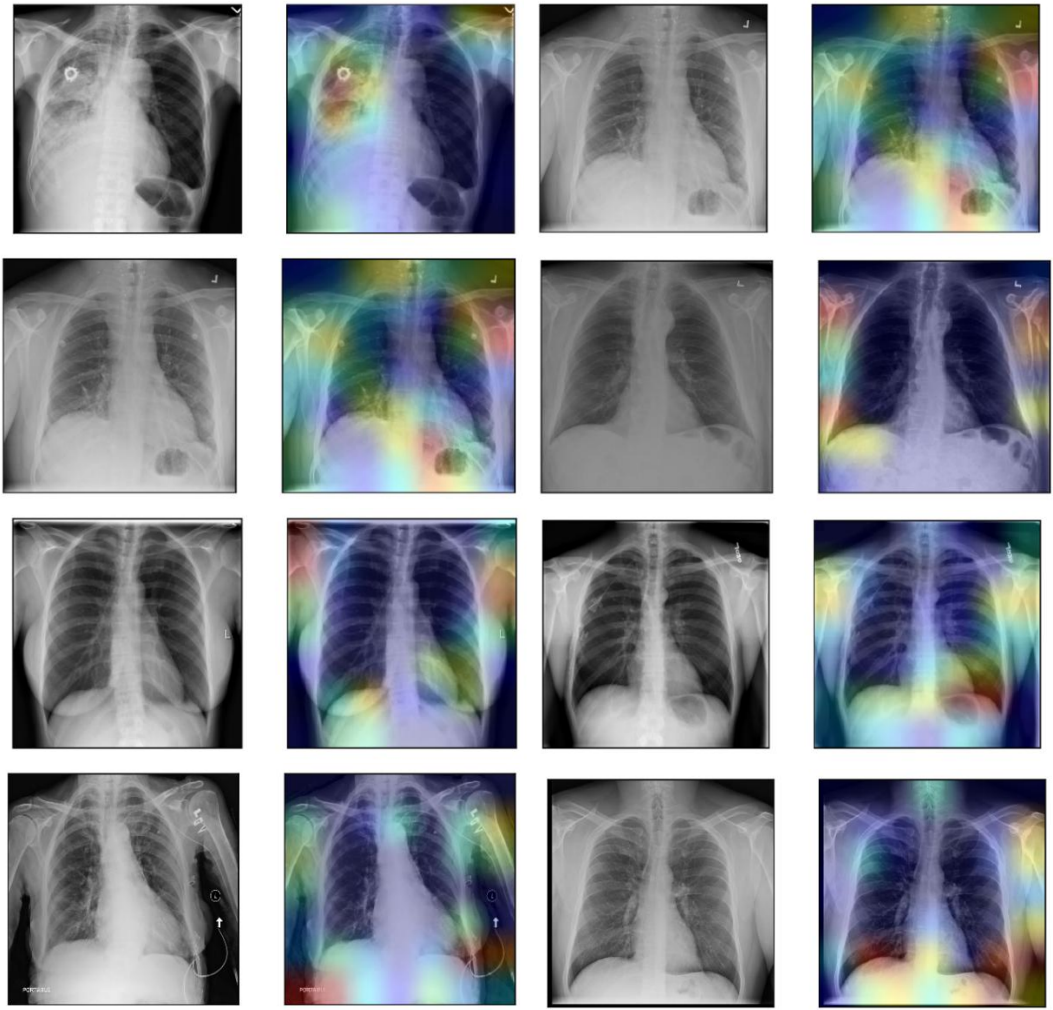

Figure 9: Heatmap visualization of unsegmented chest X-rays.

### 3.4 RQ2. Does data augmentation necessarily improve the performance of the model?

In general, data augmentation can help increase the accuracy of the machine learning models by training them on more variations and data. It acts as a regularizer that can help reduce class imbalance while training. However, in case of very small dataset, a disproportionate generation of data could lead to loss of generalization.

To test our hypothesis in RQ2, we train the models on two levels, one for augmented dataset and another for un-augmented ones. In our study, augmentation was carried out in four different ways on 600 images (300 positive and 300 negative x-rays) respectively, to obtain a total of 2400 new images. The different techniques involved tilting the X-ray by 45 degrees, scaling the image to dimensions 350x450, increasing the image contrast, and cropping of the image.

We trained and tested the models by adding 240 augmented images to each test till all 2400 images were added to the original dataset. For all models, generally, the training accuracy showed a slightly positive trend, but the increase in accuracy is almost negligible. However, test accuracy decreased consistently with an increase in augmented data. The test accuracy fell with an average loss of 24.75% across all models. SqueezeNet showed the highest loss in accuracy at 25.5% and AlexNet had the lowest test accuracy throughout all the models in all augmented image data sets, at 60.1%. Summarizing, in fig 10, we can see that test accuracy starts dropping on adding more augmented images. This indicates that the model is over fitting on the dataset and is hence showing poor test performance.

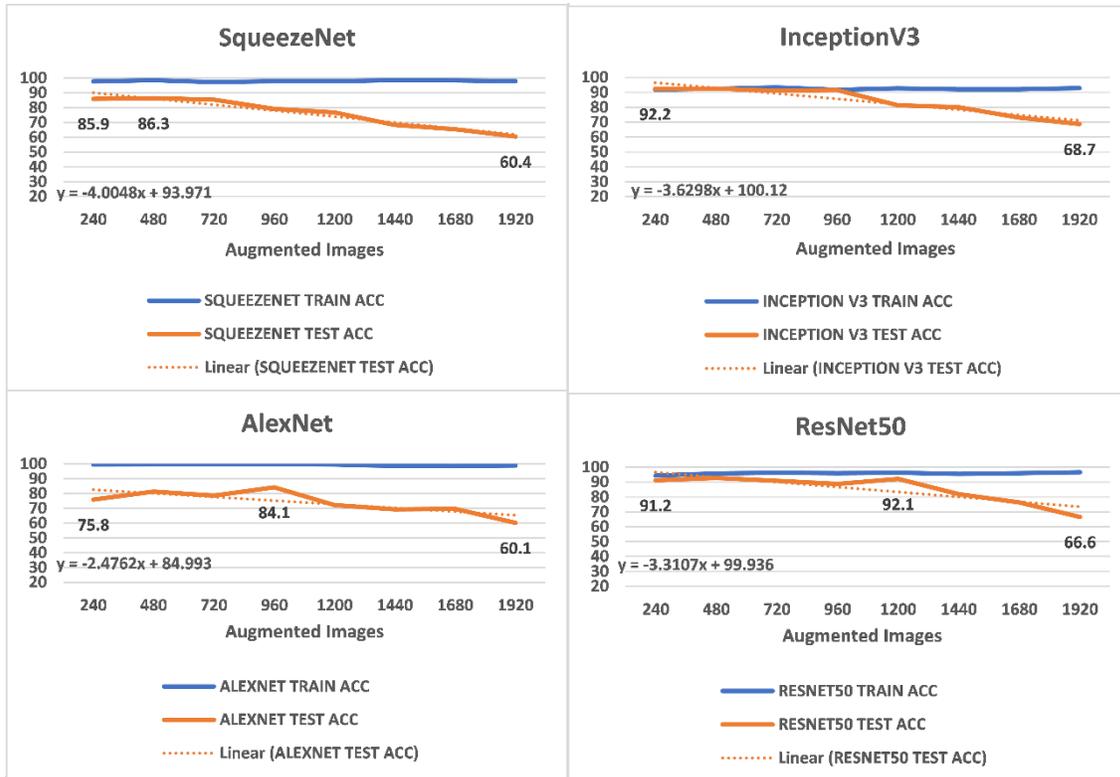

Figure 10: Performance of models in regards to percentage augmentation of total dataset.

### 3.5 Results of Proposed Methodology, SDL-COVID

As described in section 2.1 & 2.2, we have used HE for image enhancement and U-Net for lungs segmentation followed by model training as depicted in figure 4. Table 2 and 3 respectively describe the hyperparameters used for training the model and the various evaluation metrics for the trained model over the test set. Finally, figure 11 shows the confusion matrix obtained during model evaluation over the test dataset.

Table 2: Hyperparameters used for model training

| Measure | Value |
|---|---|
| Batch Size | 32 |
| Epochs | 30 |
| Learning Rate | 0.001 |
| Optimizer | Adam |
| Train Split | 3439 (80%) |
| Validation Split | 860 (20%) |

Table 3: Various evaluation metrics obtained for trained model over the test set

| Measure | Value |
|---|---|
| Sensitivity | 0.9477 |
| Specificity | 0.9500 |
| Precision | 0.9521 |
| Negative Predictive Value | 0.9455 |
| False Negative Rate | 0.0523 |
| False Discovery Rate | 0.0479 |
| False Positive Rate | 0.0500 |
| Accuracy | 0.9488 |
| F1 Score | 0.9499 |

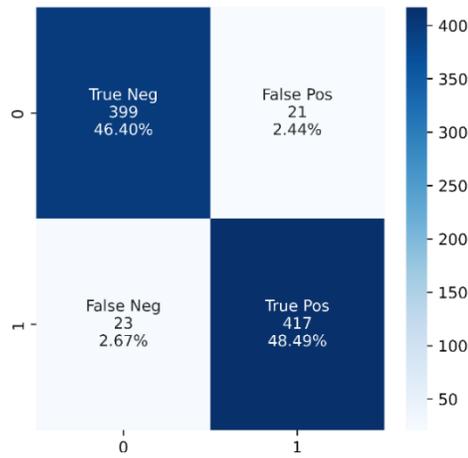

Figure 11: Confusion matrix obtained while evaluation of the proposed model on the test dataset.

### 3.6 Comparison of SDL-COVID with earlier works on segmentation

Table 4 depicts how our model has performed compared to earlier works where segmentation has been applied. Although Ahmed et al [7] achieved better results, it is to be noted that their test dataset was very small and comprised of only 91 images, possibly leading to good results. However, the small test size can also lead to loss of generalization. Rahman et al [8] also achieved decent performance, however, they used data augmentation to double the count of images. This level of augmentation does not really offer multiple datapoints and might lead to overfitting as well.

Table 4: Performance of our model compared to earlier works with segmented lungs

| Reference | Dataset | (Acc, Se, Pre, F1) | Trade-offs |
|---|---|---|---|
| [7] | +ve 410; -ve 500 | 99.26, 98.82, 98.53, 99.25 | • Test dataset too low, contains only 91 images possibly leading to loss of generalization.<br>• Performed data augmentation further leading to loss of generalization. |
| [8] | +ve 510; -ve 2160 | 96.29, 96.28, 96.29, 96.28 | • To deal with dataset imbalance, they have generated additional 50% data through augmentation – This level of augmentation might lead to overfitting and does not really offer multiple datapoints. |
| [13] | +ve 510; -ve 2160 | -, -, -, 83.00 | • Imbalanced class distribution,<br>•Used horizontal flipping for augmentation – not valid. |
| [20] | +ve 426; -ve 426 | 81.00, 76.8, -, 80.00 | • Impractical from utility point of view as accuracy drops drastically for patients having mild or moderate symptoms.<br>• Study is based on smart data, could lead to loss of generalization. |
| [22] | +ve 426; -ve 426 | 84.23, -, -, 81.31 | • Dataset small, average performance. |
| **Our work: SDL-COVID** | **+ve 2200; -ve 2100** | **94.88, 94.77, 95.21, 94.99** | - |

## 4 FURTHER INSPECTION OF RQS AND MODEL'S DECISION

In order to help radiologists, make accurate decisions, model's decision-making process must be explainable in a way that clinicians can rely on the model [34]. In this discussion, we further inspect the reliability of our findings.

An interesting observation in the case of unsegmented lungs is that, although the neural network is learning features extrinsic to the lungs, nevertheless, it is not affecting the overall performance of the model and it is able to classify covid images correctly. Upon training the models on the segmented data we find an overall decrement in train and test accuracies as compared to un-segmented lungs (fig. 12).

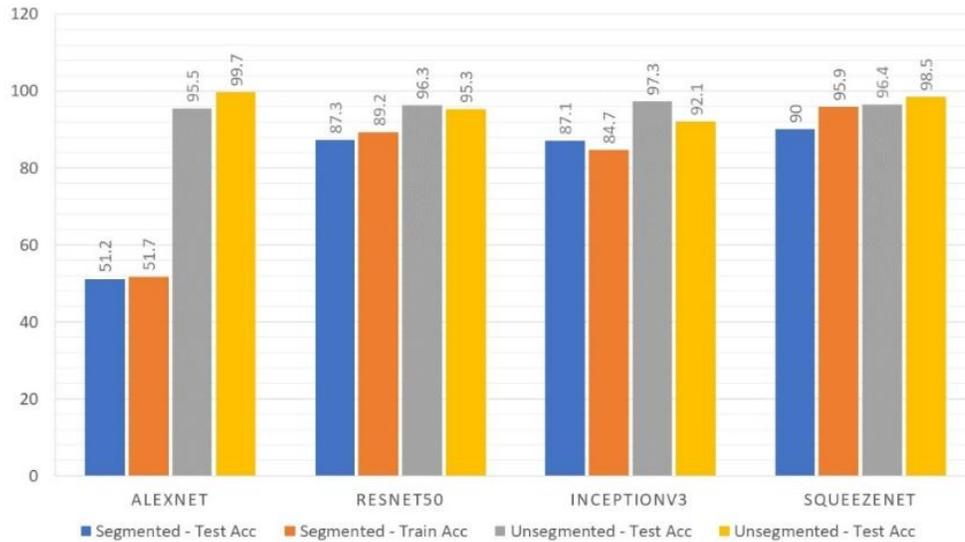

Figure 12: Comparison of accuracy of Unsegmented and Segmented results over various models.

However, an important point to be considered here is that in a non-segmented image, the extrinsic features which are although irrelevant, can still be consistently present in one set of images and thus can help the network distinguish between the covid and non covid classes. For example, most of the covid-19 X-rays are of elderly people so their x-rays might have a certain level of similarity in terms of size of the rib cage, collar bones etc. Further, the type of x-ray machine used to collect the data and the demographic of the curated dataset could also affect the learning of the model. So, even when the model is not learning relevant features from the lungs section, these additional factors (age, type of x-rays etc.) can account for correct classification of the image. Therefore, unless we do segmentation, we can't be sure of model's learning capability as was evident in the grad cam analysis (fig 9). However, when the same grad cam analysis is run over the segmented lungs (fig.13), we can clearly see that the features extracted are coming from the lungs section solely and therefore are relatively much more reliable. Hence, numerical accuracy alone cannot be the relied upon as the sole criteria for making decision in case of un-segmented lungs. However, when the same grad cam analysis is run over the segmented lungs (fig.13), we can clearly see that the features extracted are coming from the lungs section solely and therefore are relatively much more reliable. Hence, numerical accuracy alone cannot be the relied upon as the sole criteria for making decision in case of un-segmented lungs.

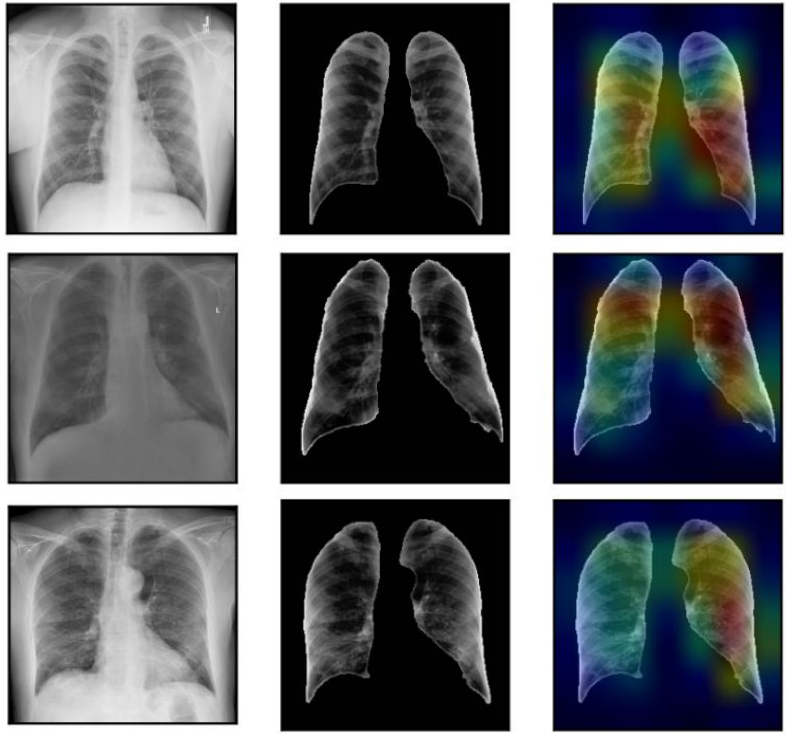

Figure 13: From left – Original CXR, HE-segmented lungs and respective heat map.

Additionally, we also provide counterfactual explanation of the predicted class by highlighting which region correspond to the opposite class through grad cam analysis (fig. 14, 15).

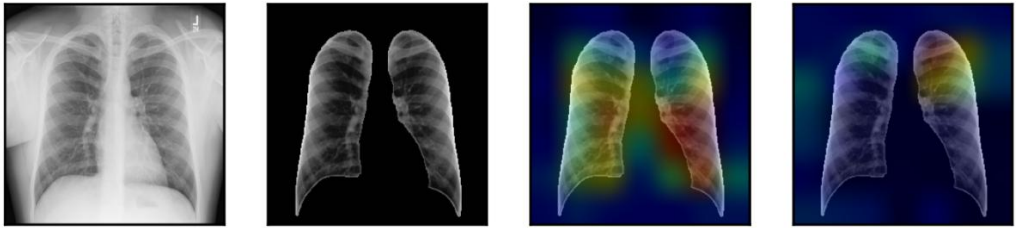

Figure 14. From left – Original CXR (Covid +ve), HE-segmented lungs, heatmap showing particular region that triggered the positive prediction and its counterfactual explanation.

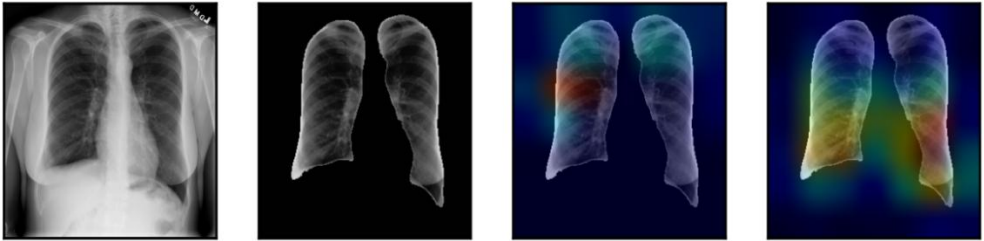

Figure 15. From left – Original CXR (Covid -ve), HE-segmented lungs, heatmap showing particular region that triggered the negative prediction and its counterfactual explanation.

It is quite evident from fig 14 and 15, that the positive and negative classification are complimentary and the pixels that trigger the positive prediction are opposite to the pixels that trigger the negative ones.

Another point to note is that in the case of medical imaging, the rate of false positive and false negative also play a critical role. Fig. 16 shows confusion matrices generated by the model over test dataset of unsegmented and segmented lungs respectively. Lower false positive rate in case of segmented lungs again supports our conclusion. Thus, although accuracy dropped after segmentation, it is advisable to diagnose the patient with segmented dataset rather with non-segmented ones.

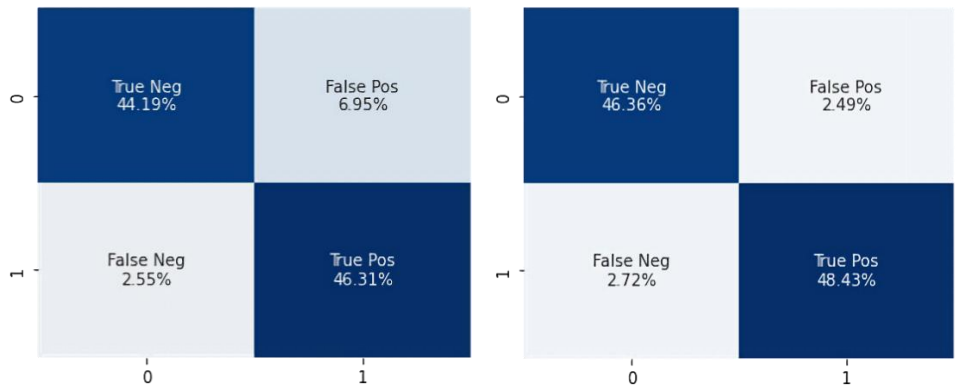

Figure 16. From left –Confusion matrix generated over test dataset of Unsegmented and segmented lungs images.

Regarding augmentation, while most of the earlier works have assumed that augmentation will give good results and directly applied augmentation to their dataset, some have not considered augmentation at all. Both these cases have an assumption based upon standard practices or theoretical backing. Our work however has a distinct advantage in the sense that we trained with and without augmentation. We then inspected the results to verify that indeed augmentation does not help the model and instead leads to overfitting.

# 5 CONCLUSION

In this study, we have presented a segmentation based deep learning methodology for reliable covid-19 X-ray classification. We establish how numerical accuracy alone cannot be the relied upon as the sole criteria for making decision in case of un-segmented lungs and unless we do segmentation, we can't be sure of model's learning capability which was also evident in the grad cam analysis.

Parallelly, we also study the effect of data augmentation on model performance. Test accuracy continuously decreased with an increase of augmented images in the dataset, on average, the test accuracy fell by 19.04% when the entire dataset was augmented. This drop in the performance indicates that the models are over fitting.

Overall, we conclude that features extracted from segmented lungs x-ray images without data augmentation are much more reliable and accurate for covid-19 classification. Despite the existence of numerous studies on covid-19 lungs classification, to the best of our knowledge there has been no systematic study to observe the impact of lungs segmentation and data augmentation in this much detail. We hope our study provides researchers a nuanced understanding of covid-19 lungs classification and eventually contribute in healthcare facilities.

## STATEMENTS AND DECLARATIONS:

- **Competing interests**: The authors declare that they have no known competing financial interests or personal relationships that could have appeared to influence the work reported in this paper.
- **Funding**: No funding was received for this project.